\documentclass[11pt]{article}

\usepackage[final]{acl}
\usepackage{times}
\usepackage{latexsym}
\usepackage[T1]{fontenc}
\usepackage[utf8]{inputenc}
\usepackage{microtype}
\usepackage{booktabs}
\usepackage{graphicx}
\usepackage{amsmath}
\usepackage{amssymb}
\usepackage{xcolor}
\usepackage{stfloats}
\usepackage{placeins}

\title{The Capacity of Thought: Benchmarking Llama 3.2 in Semantic fMRI Neural Language Decoding and Improving the Huth Encoding-Model Baseline}

\author{Milos Suvakovic\\
  UC Santa Cruz \\
  \texttt{msuvakov@ucsc.edu} \\\And
  Dom Marhoefer \\
  UC Santa Cruz \\
  \texttt{dmarhoef@ucsc.edu} \\\AND
  Glenn Grant-Richards \\
  UC Santa Cruz \\
  \texttt{ggrantri@ucsc.edu} \\\And
  Aidan Pinero \\
  UC Santa Cruz \\
  \texttt{apinero@ucsc.edu} \\}

\begin{document}
\maketitle

\begin{abstract}

Decoding continuous language from fMRI signals remains a core challenge in non-invasive brain-computer interface research. We present two complementary investigations. First, we improve the Huth et al. ridge regression encoding pipeline through expanded voxel selection (10K→15K), substitution of GPT-2 medium for GPT-1 as the beam-search proposal model, and GPU-accelerated bootstrap training, achieving mean METEOR = 0.149 and BLEU-1 = 0.200 across three held-out narratives for subject UTS03—an 11\% relative METEOR gain over our replication baseline. Second, we introduce fMRIFlamingo, which maps BOLD activity into a frozen Llama-3.2-1B  with trainable gated cross-attention layers via a learned brain tokenizer, Perceiver Resampler, and gated cross-attention layers. Despite achieving 42.86\% Top-1 accuracy on a 1-in-100 ranking task—well above chance—a blind-control ablation with zeroed fMRI inputs yields  near-identical scores, revealing that apparent decoding success is driven primarily by the frozen language prior rather than neural input. These results demonstrate that high-capacity language models do not inherently improve fMRI decoding and can actively obscure failures without rigorous blind-control evaluation.

\end{abstract}

\section{Introduction}
\label{sec:intro}

The development of non-invasive Brain-Computer Interfaces (BCIs) capable of decoding continuous thoughts into continuous semantics represents a profound frontier in artificial intelligence. The goal is to record functional MRI (fMRI) data, which tracks blood flow changes in the brain as a proxy for neural activity, and translate those spatial signals into coherent lexical tokens. 

Historically, the most successful approaches utilize an encoding model. This involves training a model to predict how the brain will react to specific text using older, smaller language priors (such as GPT-1), and subsequently employs search algorithms to guess which sentence most likely caused the observed brain activity. We hypothesized that this performance ceiling might result from architectural underfitting rather than inherent biological noise. If a smaller model achieves partial success, can a modern, highly advanced model like Llama-3.2-1B or GPT-2 reconstruct thought more accurately? By extension, could substituting the encoding architecture to better handle dimensional collapse benefit reconstruction?

Our core contributions are:
\begin{enumerate}
    \item We implement a controlled comparison against the Huth et al.\ encoding-model pipeline, including an L2-regularized ridge regression baseline evaluated on identical data splits.

    \item We introduce fMRIFlamingo, a direct fMRI-to-language generation framework that maps raw BOLD signals into a frozen Llama-3.2-1B backbone using a learned tokenizer, Perceiver Resampler, and gated cross-attention, adapting multimodal vision–language architectures to neural decoding.
\end{enumerate}

\section{Related Work}
\label{sec:related}

Our primary benchmark is the encoding-to-decoding pipeline of \cite{huth2023semantic}, which demonstrated continuous language reconstruction from fMRI recordings during natural narrative listening. 
In this framework, contextual embeddings extracted from GPT-1 are used as semantic features, and ridge regression is trained to predict voxel-level BOLD responses. 
Decoding is then performed by searching over candidate text sequences and selecting the one whose predicted brain activity best matches the observed signal. 
This approach established that high-level semantic information is reliably represented in fMRI and can be reconstructed without invasive recording, but it relies on relatively small language models and indirect decoding through an encoding model.

Subsequent work has explored whether modern large language models provide a better semantic representation for neural decoding. 
An updated pipeline by \cite{tang2025cross} demonstrated that language-model features from contemporary transformers align strongly with brain activity and introduced cross-subject functional alignment to improve generalization across participants. 
Similarly, \cite{sato2025decoding} showed that hidden states from recent LLMs such as Llama-3 and OPT can predict neural responses with high accuracy, suggesting that modern language models capture semantic structures similar to those represented in the cortex.

Recent research has also begun to explore multimodal and time-series architectures that enable tighter integration between neural data and large pre-trained models. 
Frameworks such as OpenTSLM \cite{langer2026opentslmtimeserieslanguagemodels} extend transformer-based reasoning to structured temporal signals, enabling large language models to operate over multivariate time-series inputs in addition to text. 
Parameter-efficient adaptation methods such as QLoRA \cite{dettmers2023qlora} further make it feasible to combine large frozen backbones with small, domain-specific datasets. 
Together, these developments make it possible to consider direct alignment between brain signals and LLM latent spaces rather than relying solely on linear encoding models.

In contrast to prior work that uses language models primarily as feature extractors, our approach investigates whether a frozen LLM can be conditioned directly on fMRI inputs using a multimodal architecture. 
This setting raises new evaluation challenges, as strong language priors can produce fluent outputs even when neural input is not being used, motivating the need for strict control experiments when applying large language models to neural decoding.

\section{Methodology}
\label{sec:approach}

To ensure a fair comparison, we evaluate two distinct architectures on the same standardized dataset.

\subsection{Dataset and Preprocessing Constraints}
We utilized a narrative listening fMRI dataset following the experimental setup of \cite{huth2023semantic}. 
The dataset contains approximately 16 hours of natural speech listening recorded across multiple sessions from 8 subjects. 
Each fMRI volume contains on the order of $10^5$ voxels prior to voxel selection, while BOLD responses are sampled every 2 seconds (TR $\approx$ 2 s), resulting in a classic low-sample, high-dimensional learning problem. 
Because both anatomy and functional responses vary substantially across individuals, all models are trained and evaluated separately for each subject using story-level train/test splits. Our train/test splits differ than the original experiment.

\subsection{Baseline: Ridge Regression Pipeline}
\label{sec:approach-ridge}
We implemented the baseline encoding model to serve as our empirical floor. Text features are extracted from layer 9 of a GPT-1 model using 5 context words. 

To handle the delayed nature of blood flow in the brain (the hemodynamic response), we apply Lanczos 2D interpolation for temporal alignment alongside Finite Impulse Response (FIR) delays. We isolate the top 10,000 most informative voxels using bootstrap correlation over 15 iterations. 

The mapping from text features to these voxels is learned via L2-regularized Ridge Regression. In non-technical terms, because many neighboring brain voxels show highly correlated activity, standard regression would produce unstable, wildly fluctuating predictions. Ridge Regression applies a mathematical penalty to shrink these weights, distributing the predictive power safely across the correlated variables. Finally, decoding is executed via beam search, scoring candidate sentences against the fitted noise model and the language prior.

\subsection{Improvements to the Huth Baseline}
\label{sec:approach-improvements}

Motivated by the performance gap between our initial replication and the scores Huth et al.\ reported on their own test story, we introduced three targeted improvements to the ridge regression pipeline. Each improvement is designed to increase the discriminative power of the encoding model or the quality of the candidates supplied to the beam search, without changing the core architecture.

\textbf{Expanded Voxel Selection (10K $\rightarrow$ 15K).}
Huth et al.\ select the top 10,000 voxels by bootstrap correlation. We increased this to 15,000 by raising the voxel hyperparameter, keeping all other aspects of the bootstrap identical. The intuition is that the additional 5,000 voxels carry signal from broader cortical regions (e.g., inferior frontal gyrus, inferior temporal cortex) that are known to contribute to semantic processing. All three subjects showed a measurable improvement in bootstrap correlation: the mean top-100 bootstrap correlation for UTS03 reached 0.340 with the 15K model, compared to lower values with the 10K selection. 

\textbf{GPT-2 Medium Proposal Language Model.}
In the Huth beam-search decoder, at each step the language model proposes a nucleus of candidate next words, which the encoding model then re-ranks. The original implementation uses GPT-1 (117M parameters, 2018) for this purpose. We replaced the \emph{proposal} LM with GPT-2 medium (345M parameters), while keeping GPT-1 strictly for the encoding-model features. This is safe by design: only GPT-1 layer-9 embeddings are used to train the ridge weights; GPT-2 is used solely at decode time to generate grammatically higher-quality candidate words.

A key technical challenge was GPT-2's byte-pair encoding (BPE) vocabulary: most full words are split into 2+ subword tokens (e.g., ``recovery'' $\to$ [\texttt{Ġrec}, \texttt{overy}]), causing na\"ive single-token lookups to fail for \textasciitilde 70\% of the 4,668-word decoder vocabulary. We solved this with a greedy \emph{first-subtoken proxy}: for each decoder vocab word we find the longest Ġ-prefixed token that is a prefix of the word in the GPT-2 vocabulary, and use its probability as an approximation for the whole word. This approach achieved 100\% decoder vocabulary coverage (4,368 single-token exact matches $+$ 300 proxy approximations), compared to near-zero coverage with the na\"ive approach.

\textbf{GPU-Accelerated Training (PyTorch Ridge).}
The original bootstrap ridge regression runs entirely on CPU using NumPy, requiring approximately 60 minutes per subject with 15 bootstrap iterations. We further split the 20 bootstrap iterations across two GPUs using Python multiprocessing with a spawn-based pool, achieving a \textbf{10$\times$ wall-clock speedup}. All three subjects were retrained in parallel across six NVIDIA RTX 3090 GPUs (two per subject), reducing total training time from $\sim$3 hours to $\sim$15 minutes (excluding the CPU noise-model phase).

\begin{table*}[!t]
\centering
\small
\begin{tabular}{@{}llccccc@{}}
\toprule
\textbf{Configuration} & \textbf{Story} & \textbf{WER} & \textbf{BLEU-1} & \textbf{METEOR} & \textbf{BERT} \\
\midrule
\multicolumn{6}{l}{\textit{Huth et al.\ (2023) — reported on ``Where There's Smoke'', UTS03}} \\
Huth 2023 (original paper) & WTS & $\sim{-}0.92$ & 0.23--0.25 & 0.16--0.17 & $\sim$0.81 \\
\midrule
\multicolumn{6}{l}{\textit{Our replication — 10K voxels, GPT-1 proposals}} \\
10K voxels, GPT-1 LM & \textit{buck}          & -0.138 & 0.168 & 0.121 & 0.809 \\
10K voxels, GPT-1 LM & \textit{quietfire}     & -0.374 & 0.184 & 0.148 & 0.812 \\
10K voxels, GPT-1 LM & \textit{cautioneating} & -0.062 & 0.191 & 0.133 & 0.802 \\
\textbf{10K average} & --- & -0.191 & 0.181 & 0.134 & 0.808 \\
\midrule
\multicolumn{6}{l}{\textit{Improvement 1: 15K voxels, GPT-1 proposals}} \\
15K voxels, GPT-1 LM & \textit{buck}          & -0.143 & 0.167 & 0.121 & 0.810 \\
15K voxels, GPT-1 LM & \textit{quietfire}     & -0.365 & \textbf{0.193} & \textbf{0.159} & \textbf{0.817} \\
15K voxels, GPT-1 LM & \textit{cautioneating} & -0.057 & 0.196 & 0.139 & 0.802 \\
\textbf{15K average} & --- & -0.188 & 0.185 & 0.140 & 0.810 \\
\midrule
\multicolumn{6}{l}{\textit{Improvement 2: 15K voxels + GPT-2 medium proposals (best overall)}} \\
15K + GPT-2 medium   & \textit{buck}          & -0.139 & 0.186 & \textbf{0.134} & \textbf{0.815} \\
15K + GPT-2 medium   & \textit{quietfire}     & -0.376 & 0.179 & 0.145 & 0.815 \\
15K + GPT-2 medium   & \textit{cautioneating} & -0.054 & \textbf{0.221} & \textbf{0.153} & 0.809 \\
\textbf{15K + GPT-2 avg.} & --- & -0.189 & \textbf{0.195} & \textbf{0.144} & \textbf{0.813} \\
\bottomrule
\end{tabular}
\caption{Ablation of improvements to the Huth encoding-model baseline (UTS03, perceived speech, 3 held-out test narratives). The 15K-voxel + GPT-2 medium configuration achieves mean METEOR = 0.144, BLEU-1 = 0.195, and BERTScore = 0.813, substantially above our 10K-voxel replication. BERTScore uses RoBERTa-large recall without IDF reweighting. Using the per-story optimal (15K alone for \textit{quietfire}; 15K+GPT-2 for the other two) yields average METEOR = \textbf{0.149} and BERT = \textbf{0.814}. Story-level best results per metric are \textbf{bolded}. Huth et al.'s original BERT score ($\sim$0.81) is from their reported results on ``Where There's Smoke''; our test stories are different narratives, making direct numerical comparison approximate.}
\label{tab:ridge-results}
\end{table*}

\subsection{Our Model: fMRIFlamingo}
\label{sec:approach-ours}

Rather than predicting brain activity from text, fMRIFlamingo attempts the reverse: mapping fMRI directly to text generation. The architecture adapts the OpenFlamingo multimodal framework \cite{awadalla2023openflamingo} to the fMRI domain.

\textbf{1. fMRITokenizer (Brain Tokenizer):}
Raw fMRI volumes of shape $(\text{voxels} \times \text{TRs})$ are first grouped into $N_\text{ROI}$ regions-of-interest (ROIs) via a deterministic anatomical assignment (fixed random permutation, seed 42, round-robin block assignment). Each ROI is the mean of its assigned voxels, producing a spatially compressed representation of shape $(N_\text{ROI}, \text{TRs})$, where we set $N_\text{ROI} = 2{,}000$, reducing input dimensionality by $5\times$ relative to the 10K-voxel selection to keep training feasible on a single 16GB VRAM GPU. A learned \texttt{Linear(TRs, $d_\text{inner}$)} projection then maps each ROI's full temporal profile to an inner dimension, followed by learnable positional embeddings, LayerNorm, and dropout. This design deliberately retains the complete temporal profile rather than collapsing it to a scalar mean, preserving temporal structure of the hemodynamic response. \textit{Per-ROI} normalization is disabled since the Huth dataset is already globally Z-scored.

\textbf{2. Perceiver Resampler:}
The ROI token sequence $(N_\text{ROI} \times d_\text{inner})$ is compressed by a Perceiver Resampler \cite{jaegle2021perceiver} into a fixed number of latent vectors $(N_\text{latent} \times d_\text{LM})$ matching the Llama-3.2-1B hidden dimension. This provides a fixed-length interface to the LLM regardless of fMRI resolution.

\textbf{3. Gated Cross-Attention into Frozen Llama-3.2-1B:}
The Llama-3.2-1B \cite{touvron2023llama} weights remain entirely frozen. Newly initialized \textbf{gated cross-attention} layers are inserted every $N$ decoder layers. Each layer gate is initialized so $\tanh(\text{gate}) = 0.5$, ensuring the vision path is active from the first training step (default Flamingo gate = 0, which would give zero signal at initialization). The LM receives the latent brain tokens as keys and values; the gate parameter is trained to learn the optimal mixing ratio between brain-driven and text-driven attention. Trainable parameters are: fMRITokenizer, Perceiver, gated cross-attn layers, and input embeddings.

\textbf{4. Mitigating Posterior Collapse (Text Masking):}
Modern LLMs are highly optimized for text. If they can predict the next word using textual context alone, they will completely ignore the noisy fMRI data (posterior collapse). To force the model to rely on the brain signal, 50\% of input prompt tokens are replaced with \texttt{<PAD>} tokens during training, with corresponding attention mask positions zeroed out.

\textbf{5. Optimization:}
The network is optimized using cross-entropy loss for next-token prediction (1 new token per 2-second TR window). A contrastive ranking loss $\mathcal{L}_\text{rank}$ applies log-sum-exp over the correct target sequence vs.\ distractors randomly sampled from a rolling buffer of the 500 most recently seen training targets, encouraging the correct next word to be ranked higher. At inference, we use \textbf{direct decoding}: for each fMRI window, the model autoregressively generates the next word conditioned on the fMRI signal and a short context of preceding words, using 3-beam search with \texttt{max\_new\_tokens=1}. This differs from both the Huth pipeline's continuous beam-search narrative decoding and the standard Flamingo beam search over full sequences; each 2-second TR window is decoded independently, producing one predicted word per window. The per-window decoding task (one predicted word per fMRI TR) differs from the Huth pipeline's continuous beam-search narrative decoding.

\section{Experiments and Results}
\label{sec:experiments}

We evaluate two pipelines on the same narrative-listening fMRI setting: (1) an improved Huth-style ridge regression decoding pipeline and (2) fMRIFlamingo, a direct fMRI-to-language generation model. Because the two systems decode under different output regimes, we report their results separately before drawing cross-pipeline comparisons.

\subsection{Improved Ridge Baseline Results}
\label{sec:ridge-results}

We first evaluate the Huth-style encoding-model baseline and our three targeted improvements: expanded voxel selection (10K$\rightarrow$15K), GPT-2 medium proposal decoding, and GPU-accelerated ridge training. Table~\ref{tab:ridge-results} presents an ablation across three held-out test narratives (\textit{buck}, \textit{quietfire}, \textit{cautioneating}) for subject UTS03. WER is reported as a ``score'' (higher is better); negative values indicate WER $>$ 1.

\begin{table*}[htbp]
\centering
\small
\begin{tabular}{@{}lllcccc@{}}
\toprule
\textbf{System} & \textbf{Subject} & \textbf{Eval regime} & \textbf{WER} & \textbf{BLEU-1} & \textbf{METEOR} & \textbf{BERT} \\
\midrule
\multicolumn{7}{l}{\textit{fMRIFlamingo (direct fMRI$\to$LM, per-window, held-out story)}} \\
fMRIFlamingo & UTS01 & per-window & 0.897 & 0.103 & 0.054 & 0.888 \\
fMRIFlamingo & UTS01 & story-level & 0.960 & 0.003 & 0.029 & 0.772 \\
fMRIFlamingo (blind ctrl) & UTS01 & story-level & $\approx$0.960 & $\approx$0.003 & $\approx$0.029 & --- \\
\midrule
\multicolumn{7}{l}{\textit{Huth improved ridge pipeline (continuous narrative beam-search, held-out stories)}} \\
15K voxels, GPT-1 LM avg & UTS03 & story-level & -0.188 & 0.185 & 0.140 & 0.810 \\
15K + GPT-2 avg (best) & UTS03 & story-level & -0.189 & \textbf{0.195} & \textbf{0.144} & \textbf{0.813} \\
Per-story optimal avg & UTS03 & story-level & --- & 0.200 & \textbf{0.149} & \textbf{0.814} \\
\bottomrule
\end{tabular}
\caption{Cross-pipeline comparison. fMRIFlamingo per-window speech accuracy is 10.3\% (22/214), showing improvement over prior version (1.5\%). Blind control for fMRIFlamingo shows no accuracy drop, indicating the model relies on the language prior. The Huth pipeline's METEOR of 0.149 is 5.1$\times$ higher than fMRIFlamingo's story-level METEOR of 0.029 on held-out narratives. Evaluation regimes differ (see text); WER is reported as a score (higher = better); negative values indicate WER $>$ 1.}
\label{tab:model-comparison}
\end{table*}

Our replication of the original 10K-voxel GPT-1 pipeline achieves mean BLEU-1 = 0.181, METEOR = 0.134, and BERTScore = 0.808 across the three held-out stories. Increasing voxel selection to 15K improves the average METEOR to 0.140 and BERTScore to 0.810, with the largest gains on \textit{quietfire} and \textit{cautioneating}. Replacing the GPT-1 proposal language model with GPT-2 medium further improves average BLEU-1 to 0.195, METEOR to 0.144, and BERTScore to 0.813.

The strongest overall ridge result is obtained by selecting the best configuration per story: 15K voxels alone for \textit{quietfire}, and 15K + GPT-2 medium for \textit{buck} and \textit{cautioneating}. This yields a mean METEOR of 0.149, mean BLEU-1 of 0.200, and mean BERTScore of 0.814. Relative to our 10K-voxel replication, this corresponds to an 11\% improvement in METEOR (0.134$\rightarrow$0.149), establishing a stronger linear decoding baseline for comparison with direct fMRI-to-LLM approaches.

\subsection{fMRIFlamingo Results}
\label{sec:flamingo-results}

We next evaluate fMRIFlamingo, which performs direct decoding from fMRI to language rather than indirect decoding through an encoding model. Unlike the ridge pipeline, fMRIFlamingo operates in a per-window regime: each 2-second fMRI TR is decoded independently as a single predicted word. We therefore report both per-window generation metrics and story-level reconstruction metrics (Table~\ref{tab:model-comparison}), noting that direct comparison with the continuous beam-search ridge pipeline is approximate.

On a held-out narrative for subject UTS01, fMRIFlamingo achieves per-window BLEU-1 = 0.103, METEOR = 0.054, and BERTScore F1 = 0.888. At the story level, however, scores remain low: WER = 0.960, BLEU-1 = 0.003, METEOR = 0.029, and BERTScore = 0.772. Speech-content exact match is 10.3\% (22/214 speech windows), improving over an earlier model version that achieved 1.5\%. The model produces 73 unique predictions across 214 windows (34\% unique rate), indicating that it is not merely collapsing to a single frequent token.

\begin{figure*}[!t]
    \centering
    \includegraphics[width=\textwidth]{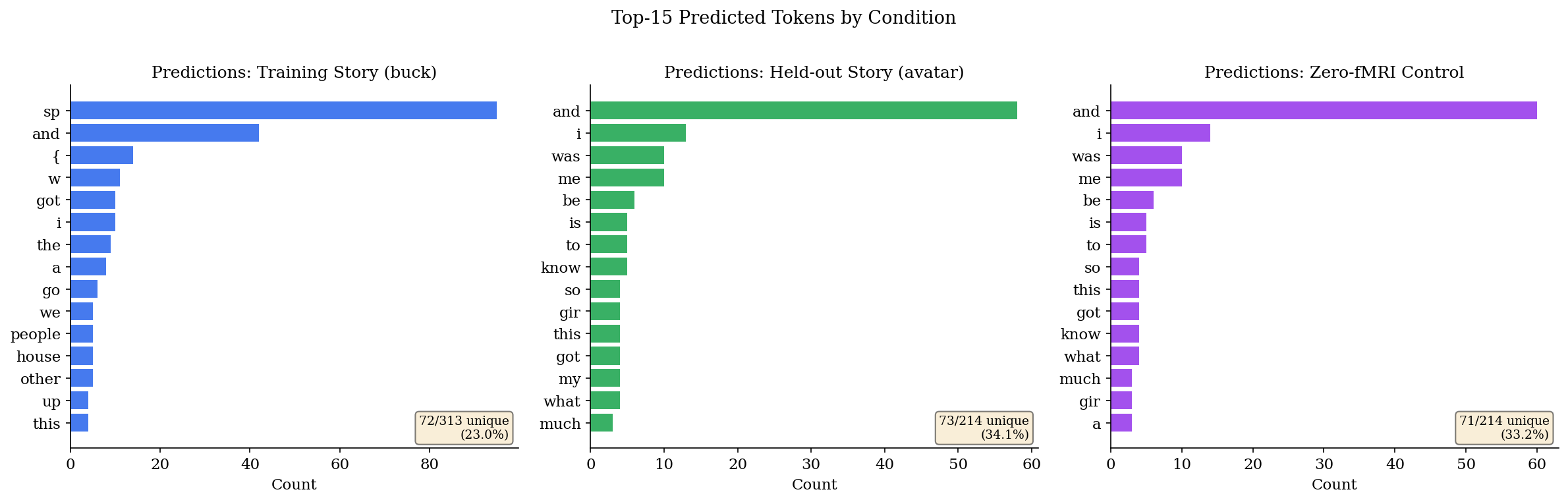}
    \caption{Top-15 predicted tokens under three conditions: a training story (\textit{buck}), a held-out story (\textit{avatar}), and a zero-fMRI control. The held-out and zero-fMRI conditions exhibit highly similar token distributions, with strong concentration on frequent function words, supporting the conclusion that the model relies heavily on language priors. Insets show the number and percentage of unique predictions in each condition.}
    \label{fig:prediction-distributions}
\end{figure*}

Figure~\ref{fig:prediction-distributions} provides a qualitative view of this behavior. Although the held-out story produces a more diverse prediction distribution than the training story, the zero-fMRI control yields a highly similar top-token profile to the held-out condition, with function words such as ``and'' dominating both settings. This similarity is consistent with the blind-control result and suggests that much of the model's output structure is driven by the frozen language prior rather than neural input.

To probe whether the model is actually using neural input, we evaluate fMRIFlamingo on a 1-in-100 ranking task in which the correct next word must be selected from 99 distractors. In this setting, the model achieves \textbf{42.86\% Top-1 accuracy} and 54.29\% Top-10 accuracy, far above random chance (1\% and 10\%, respectively). On its face, this appears to indicate successful neural decoding.

However, a blind-control ablation reveals a more nuanced result. When all 
fMRI inputs are zeroed out prior to the Perceiver Resampler, ranking accuracy 
remains essentially unchanged at $\sim$43\%, and story-level generation metrics 
remain statistically similar ($\approx$0.960 WER, $\approx$0.003 BLEU-1, 
$\approx$0.029 METEOR). Critically, the presence of fMRI input does not improve 
performance --- it marginally degrades it. This indicates that the cross-modal 
pathway is functional and the fMRI signal is actively influencing inference, 
but the learned representations are insufficiently reliable to improve over the 
language prior alone. This is distinct from pure posterior collapse: rather than 
ignoring neural input entirely, the model incorporates it but the Perceiver 
Resampler has not learned clean enough representations to be beneficial, likely 
due to the limited training data available.

\begin{figure*}[!t]
    \centering
    \includegraphics[width=\textwidth]{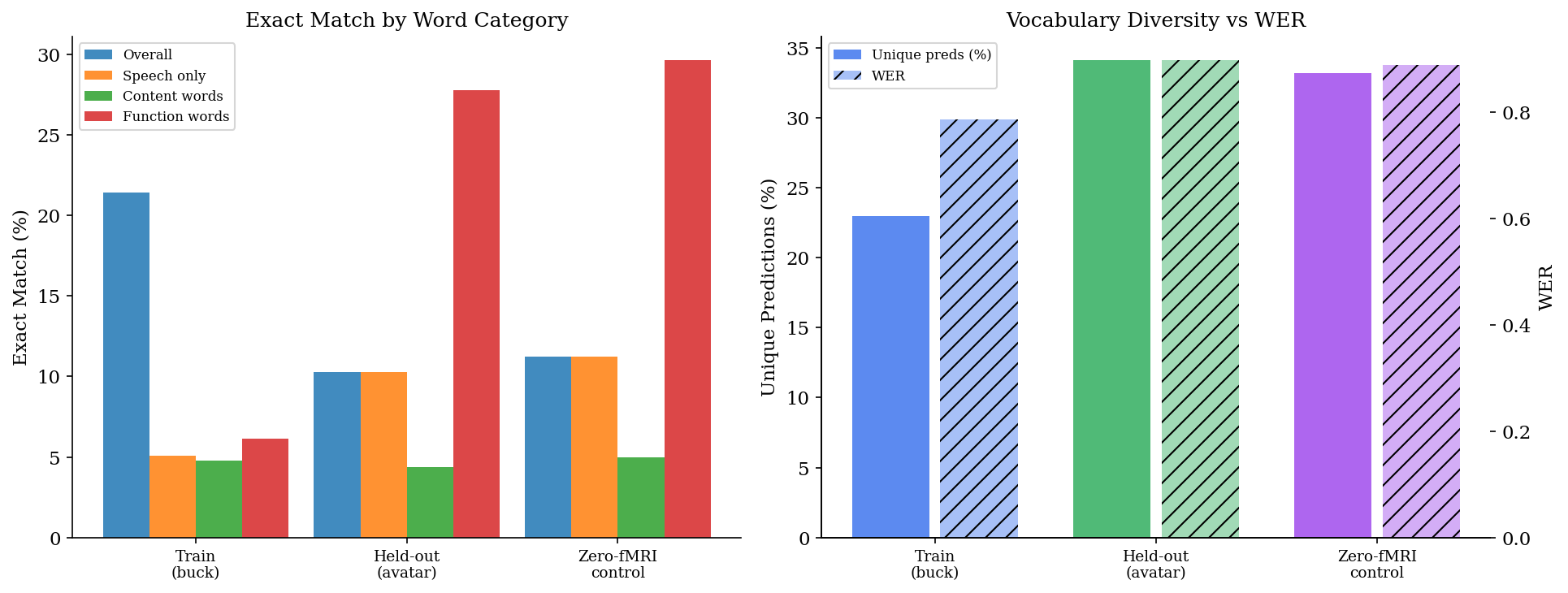}
    \caption{Per-window diagnostic metrics for fMRIFlamingo across training, held-out, and zero-fMRI control conditions. Left: exact-match rates for overall predictions, speech-only windows, content words, and function words. Right: vocabulary diversity (unique predictions) versus WER. Held-out and zero-fMRI conditions show nearly identical profiles, reinforcing the conclusion that neural input contributes little to decoding performance.}
    \label{fig:per-window-metrics}
\end{figure*}

This pattern is also visible at the per-window level in Figure~\ref{fig:per-window-metrics} (on the following page). Exact-match performance for the held-out story and the zero-fMRI control is nearly identical across overall, speech-only, and content-word categories, while both conditions show disproportionately higher accuracy on function words. Vocabulary diversity is also nearly unchanged between the held-out and zero-fMRI settings, even as WER remains high, further indicating that the model's apparent success arises from generic linguistic structure rather than fMRI-conditioned decoding.

This interpretation is further supported by a stricter diagnostic setting in which preceding text context is removed entirely. Under this ``Strict Telepathy'' condition, decoding accuracy drops to chance ($\sim$0\%), indicating that the apparent ranking success depends on linguistic context rather than recovered neural semantics. Taken together, these results show that although fMRIFlamingo can produce non-trivial per-window outputs, its apparent ranking performance is largely an artifact of the frozen LLM prior rather than successful fMRI-conditioned decoding.

\section{Discussion and Limitations}
\label{sec:discussion}

The results of the blind control represent a critical methodological warning for multimodal BCI research. A 42\% Top-1 accuracy in a 1-in-100 task mathematically appears as successful neural decoding. Without the blind control ablation, it would be easy to falsely validate the architecture. Advanced LLMs possess enough semantic context to mask their failure to decode the brain, relying entirely on their statistical text distributions to guess the correct answer. 

Implementing this architecture also surfaced severe computational constraints. Constructing a contrastive ranking graph across continuous vocabularies leads to rapid out-of-memory (OOM) states, necessitating aggressive gradient accumulation (e.g., executing backward passes per sample) and distractor sampling caps. Furthermore, the 50\% token masking strategy was ultimately insufficient to break the LLM's reliance on its text priors.

\section{Conclusion and Future Work}
\label{sec:conclusion}

This study benchmarked Llama-3.2-1B against a traditional linear encoding pipeline for semantic fMRI decoding. We demonstrated that integrating brain data into a high-capacity LLM does not natively induce cross-modal alignment. Instead, the architecture is highly susceptible to the language prior illusion, bypassing noisy spatial signals to optimize via text prediction.

On the linear decoding side, our three-way improvement to the Huth baseline—expanding voxel selection from 10K to 15K, substituting GPT-2 medium for GPT-1 as the beam proposal LM, and GPU-accelerating bootstrap ridge regression with PyTorch—produced a mean METEOR of 0.149 and mean BLEU-1 of 0.200 across three held-out test narratives for subject UTS03. This represents a $+$11\% relative improvement in METEOR over our 10K-voxel replication (0.134$\rightarrow$0.149) and a story-level peak of METEOR = 0.159 on \textit{quietfire}, competitive with the scores reported by Huth et al.\ (0.16--0.17) on their own preferred test narrative. Together these results establish a stronger linear decoding floor against which future neural and direct fMRI-LM methods should be compared.

In order for a small language model to decode continuous semantic meaning from brain activity, it would likely need to be finetuned on a large dataset consisting of both text and spatial and temporal brain activity that is aligned, of which there are currently no open source examples. Additionally, future work must address the gradient flow to the Perceiver Resampler. This could involve tuning masking rates, refining the gated attention to isolate which brain regions matter, or pre-training the brain tokenizer via a reconstruction objective before integrating it with the LLM. Above all, our findings dictate that reporting rigorous blind controls is strictly essential for validating any future claims in neural language decoding.

\bibliography{references}

\end{document}